# Foundation Models for Autonomous Robots in Unstructured Environments

Hossein Naderi[1]; Alireza Shojaei, Ph.D.[2]; and Lifu Huang, Ph.D.[3]

[1]Ph.D. student, Dept. of Building Construction, Myers-Lawson School of Construction, Virginia Tech, Blacksburg, VA. ORCID: https://orcid.org/0000-0002-6625-1326. Email: Hnaderi@vt.edu
[2]Assistant Professor, Dept. of Building Construction, Myers-Lawson School of Construction, Virginia Tech, Blacksburg, VA (corresponding author). ORCID: https://orcid.org/0000-0003-3970-0541. Email: Shojaei@vt.edu
[3]Assistant Professor, Computer Science Department, Virginia Tech, Blacksburg, VA. Email: lifuh@vt.edu

## Abstract

Automating activities through robots in unstructured environments, such as construction sites, has been a long-standing desire. However, the high degree of unpredictable events in these settings has resulted in far less adoption compared to more structured settings, such as manufacturing, where robots can be hard-coded or trained on narrowly defined datasets. Recently, pretrained foundation models, such as Large Language Models (LLMs), have demonstrated superior generalization capabilities by providing zero-shot solutions for problems do not present in the training data, proposing them as a potential solution for introducing robots to unstructured environments. To this end, this study investigates potential opportunities and challenges of pretrained foundation models from a multi-dimensional perspective. The study systematically reviews application of foundation models in two field of robotic and unstructured environment and then synthesized them with deliberative acting theory. Findings showed that linguistic capabilities of LLMs have been utilized more than other features for improving perception in human-robot interactions. On the other hand, findings showed that the use of LLMs demonstrated more applications in project management and safety in construction, and natural hazard detection in disaster management. Synthesizing these findings, we located the current state-of-the-art in this field on a five-level scale of automation, placing them at conditional automation. This assessment was then used to envision future scenarios, challenges, and solutions toward autonomous safe unstructured environments. Our study can be seen as a benchmark to track our progress toward that future.

## 1. Introduction

Automating tasks using robot capabilities has long been a desire in many unstructured and dynamic environments, such as construction sites [1]. Tasks in these environments are often diverse, and require significant human labor, physically demanding, and mostly dangerous. However, the integration of robots in these unpredictable settings remains considerably less advanced compared to their widespread use in structured environments like manufacturing [2]. This discrepancy is attributed to the complex and ever-changing nature of unstructured spaces which requires reactive and proactive behavior of robots for automating very diverse tasks [3].

Moreover, experts in robotics in unstructured environments believe that technologies are not mature enough to solve these challenges [4].

To achieve a degree of autonomy in unstructured environment, there are generally two broad solution categories for robotics: (1) pre-programming robots for specific scenarios; (2) tele-operating robots to leverage human cognitive abilities [5]. The first category employs Machine Learning paradigms, such as reinforcement learning [6] and deep learning [7], to automate specific labor-intensive and repetitive tasks [8], [9]. While these robots can deliver satisfactory precision in designated tasks, their adaptability and generalizability are often limited due to training on narrowly focused datasets designed for specific tasks. Consequently, manual adjustments may be necessary to accommodate even minor task variations [10]. In contrast, the second category involves tele-operated robots, which can be remotely operated by experts, allowing them to adapt to various tasks without the need for manual reprogramming. However, their dependency on human operators has limited their performance and productivity. For example, even slight connection delays can significantly impede robot performance in unstructured environments [11].

Foundation models are defined as large-scale Artificial Intelligence (AI) models trained on an extensive and internet-scale datasets, capable of generalizing knowledge across a wide range of tasks. Models from this family, such as such as GPT-3 [12], GPT-4 [13] Llama-2 [14] Gemini [15] Claude [16] have significantly advanced the fields of natural language processing (NLP) and computer vision, enabling machines to understand and generate human-like text and to recognize, interpret, and generate images with high accuracy [17]. Such capabilities and strengths have drawn the attention of researchers in diverse domains, ranging from medical field [18] to robotics [19]. By leveraging the multifaceted capabilities of these models, including their adaptability and ability to process multi-modal data (text, image, sound), we can explore how they offer viable solutions to the intricate problems faced in performing tasks autonomously in unstructured environments, such as construction sites and disaster zones.

Consequently, this study aims to systematically capture the current state of foundation model applications in (1) robotics and unstructured environments; (2) synthesize findings and map the current state-of-the-art in integrating robots with foundation models for unstructured environments; (3) synthesize findings and outline future potential scenarios, challenges, and associated steps toward safe, fully automated robots in unstructured environments. This study contributes to the field by: (1) providing, to the best of the authors' knowledge, the first systematic exploration of the current state of the art in the integration of foundation models and robotics in unstructured environments; (2) introducing future scenarios and pathways toward fully autonomous robots in unstructured environments; (3) serving as a benchmark for other researchers to track and follow the progress toward future safe and fully autonomous robots.

## 2. Robots in unstructured environment
### 2.1- Unstructured Environment: Study Scope

An unstructured environment can be defined as a setting characterized by a notable lack of clear organization and predefined parameters, thereby presenting a higher degree of uncertainty and complexity [20], [21]. Such environments often involve a variety of unanticipated obstacles, objects, or situations, which makes the adoption of robots particularly challenging [22], [23]. Considering these features, a wide range of settings and workspaces can be categorized as unstructured environments, thus necessitating a clarification of the boundaries for this study.

Figure 1 illustrates the scope of this study, focusing on unstructured environments to define the parameters of our exploration. Unstructured environments are marked by a wide array of tasks, limited simulation capabilities, and highly dynamic conditions, posing challenges for testing robotic applications in real-world settings. In contrast, structured environments offer well-defined parameters for robot deployment, featuring constant or low-dynamic conditions, and moderate difficulty in testing real-world scenarios, with manageable constraints. Tasks in structured environments typically exhibit lower to medium diversity, which facilitates simulation [6], [23], [24]. Furthermore, there are other environments beyond the scope of this study that are extremely harsh and remain relatively unexplored, including outer space, undersea, and deep forests. These environments are characterized by extreme conditions such as extreme pressures, temperatures, darkness, and slow communication. These factors significantly impact the tasks that robots must undertake in these environments, necessitating specific configurations to operate effectively under such extreme conditions [25]. For example, robots designed for deep-sea exploration must be able to withstand the crushing pressures of the ocean depths while performing tasks like underwater welding or biological sampling, requiring unique adaptations that is out of the scope of this study.

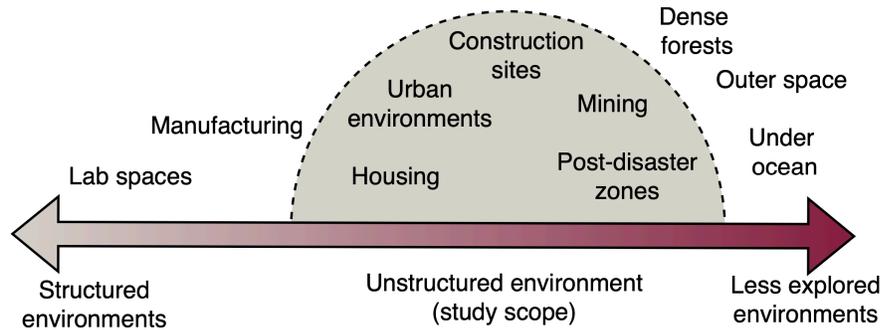

Figure 1 The study scope for unstructured environments

While all environments within the study scope are unstructured, their levels of environmental complexity vary and directly impact the requirements for robots to perform tasks effectively. To assess the environmental complexity for autonomous robots, a framework [3] suggests three primary factors: (1) the dynamics of the environment; (2) observability; and (3) uncertainty. The dynamics of the environment refer to the frequency and degree of changes and unpredictable events occurring within it. Observability involves the amount of permanent information available to an actor or robot within the environment. Uncertainty measures the degree and impact of unpredictable events within the environment. Table 1 provides an examination of unstructured environments within the scope of this study through the lens of these factors and based on rule of thumb. While the designations are applicable in most cases, the authors acknowledge some variations happening in edge cases that deviate from typical situations and require a separate study.

*Table 1* Degree of complexity in unstructured environments

| Unstructured Environments | Degree of Dynamics | Degree of Observability | Degree of Uncertainty |
|---|---|---|---|
| Housing | Medium | High | Medium |
| Urban Environments | Very high | High | Medium |
| Construction Sites | High | Low | High |
| Mining | High | Low | High |
| Post-Disaster Zones | High | Very low | Very high |

## 2.2- Theory: Deliberation for Autonomous Robots

It is imperative to understand the fundamentals and desired behaviors of autonomous robots in unstructured environments to effectively incorporate robots with them. Ghallab and et al. [3] explored this subject by attributing "deliberative acting" to robots, where "action" refers to activities undertaken by an agent or robot to effect changes in both its environment and its own state. Deliberation, within the context of action-taking, involves making decisions about which actions to pursue and how to execute them to achieve specific objectives. This cognitive process enables the robot to perceive, plan, and determine appropriate actions, combining multiple actions that collectively contribute to achieving the desired goal.

Before delving into the deliberative acting process of robots, it is essential to examine the origins of this cognitive process in unstructured environments. The need for deliberate action stems from two primary desires in unstructured environments:

(a) Autonomy: This refers to the necessity for the robot to perform its intended task independently, without human intervention.

(b) Generalizability: This pertains to the various environments in which robot can operate, and the range of tasks the robot can perform.

These factors are critical in satisfying the demand for robots in unstructured environments. In scenarios where autonomy is absent, devices that are directly operated or teleoperated do not typically require deliberation. Instead, these devices extend the capabilities of human operators who are responsible for understanding and decision-making, often with the assistance of advisory and planning tools. Such scenarios are common in applications like teleoperation or operating heavy machinery, such as excavators in construction sites, mining sites, and post-disaster zones. Conversely, an autonomous system may not require generalizability if it operates exclusively in a well-defined environment or is designed for very specific tasks. For example, manufacturing robots autonomously execute tasks like painting and welding with minimal generalizability, tailored to their specific environments.

### 2.3- Autonomous Robot Components in Unstructured Environment

To achieve a satisfactory level of deliberation for autonomous robots in unstructured environments, certain components and tasks must be thoroughly addressed. Figure 2 illustrates a conceptual view of a robot operating within such an environment. Autonomous robots are comprised of two main platforms: (1) the deliberation platform; and (2) the execution platform. The execution platform, which is influenced by the robot's morphology, includes various actuators, motors, sensors, end effectors, and manipulators. Developments in this platform are beyond the scope of this study, as our primary focus is on the deliberation platform. This platform is responsible for receiving objectives and percepts (mostly from various sensors), processing them, and generating actionable commands or communication signals.

The deliberation platform employs two main modules: (1) the cognitive module, which is responsible for all cognitive processes in robots; and (2) the acting module, which translates cognitive outputs into actionable commands. These two modules consist of interconnected components essential for the performance of the deliberation platform. Reasoning represents the highest-level cognitive process, involving the inference of new information or conclusions from existing percepts and signals. In mid-level cognitive processes, planning is crucial and involves generating a sequence of decisions to achieve specific goals or objectives. Decision-making entails choosing actions or behaviors from a set of alternatives based on available percepts and reasoning, guided by predefined criteria and objectives. Human-robot interaction (HRI) facilitates interactions with humans or other robots through verbal or non-verbal communication, including tasks such as speech recognition, natural language processing, generating spoken or written responses, and understanding gestures or facial expressions. Additionally, the perception component involves processing, perceiving, and interpreting information from the environment and communicating it with other components. This may include tasks such as object recognition, scene understanding, Simultaneous Localization and Mapping (SLAM), gesture and action recognition, depth estimation, and segmentation.

The acting module also includes components critical for its function. Robots require mechanisms for controlling their actuators (e.g., motors, joints, grippers) to effectively execute planned actions (control component). Navigation involves the robot's movement through its environment, which includes path planning, obstacle avoidance, and localization. Manipulation tasks involve physical interaction with objects in the environment, such as picking up, moving, and manipulating objects.

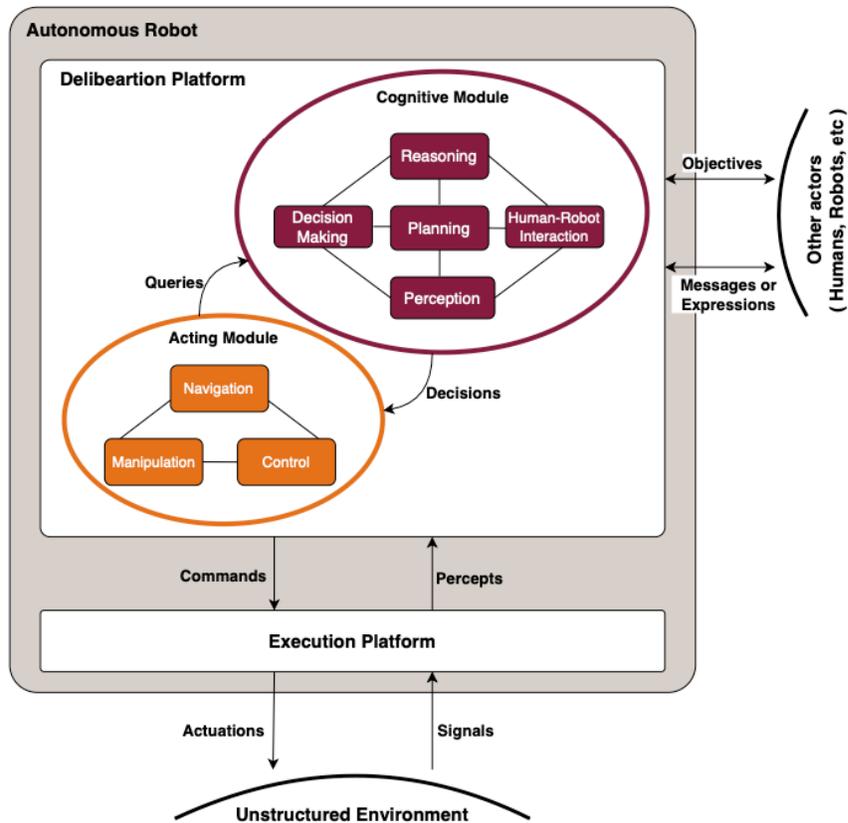

*Figure 2* Conceptual view of autonomous robots in unstructured environment (adapted from [3])

## 3. Foundation Models: Overview

### 3.1- Definition

The foundation model term is first introduced by researchers at Stanford university [26]. Unlike previous specialized AI models that were trained on task-specific data to perform specific tasks, these very large neural networks are trained on a broad and massive scale using internet-scale unlabeled data in an unsupervised manner. This training approach makes them adaptable to a wide range of general downstream tasks [27]. They are termed "foundation" models because they can serve as a starting point for developers to build upon for new tasks, instead of training a model from scratch [28], thereby representing a shift toward more generalized, adaptable, and scalable AI solutions. These models enhance their general learned information to be suited for specific tasks through mechanisms such as self-supervised learning and fine-tuning [29].

Foundation models are characterized by three main features [30]: (1) In-context learning, which enables the accomplishment of new tasks with just a few examples, without the need for retraining or fine-tuning; (2) Scaling laws, which allow for continued performance improvements as data, computational resources, and model sizes increase; (3) Homogenization, which allows certain foundation model architectures to handle diverse modalities in a unified manner. Despite all these advantages, training models at this scale incurs significant expenses. It requires considerable computational resources, necessitating specialized hardware such as GPUs or TPUs, along with the essential software and infrastructure for model training. These factors together entail substantial financial investments. Additionally, training a foundation model is time-consuming, which poses additional costs [31], [32].

### 3.2- Concept Clarification

The terms Large Language Models (LLMs) and Generative AI are commonly used interchangeably with Foundation Models in studies relevant to this domain. However, there are significant differences between them, making it necessary to clarify these distinctions to accurately determine whether a topic falls within or outside our scope. Figure 3 represents the boundaries and domains of these terms. AI, being the broadest domain, acts as the overarching umbrella that encompasses all these terms and enables machines to simulate human intelligence. Within AI, machine learning is a major subdomain that employs various algorithms and techniques to draw inferences from data patterns. A more specialized subdomain is deep learning, which utilizes multi-layer neural networks to extract features from data [33]. Considering the definition of foundation models provided in the previous section, they fall under the deep learning subdomain. Within this domain, foundation models can be fine-tuned for specific tasks. For example, LLMs [34] are specialized foundation models fine-tuned for understanding and generating human text and language. Additionally, we have Large Vision Models (LVMs), Large Audio Models (LAMs), and some emerging models, such as Large Robotic Models (LRMs) and Large Multimodal Models (LMMs) [35], which will be discussed in more detail in the following sections.

There is no clear consensus on the boundaries between Generative AI and other related terms. Generative AI is categorized based on its purpose rather than its architecture [36], focusing on AI models and tools primarily aimed at generating new content, such as text, images, videos, music, or code [37], [38]. However, not all foundation models are solely focused on content generation; they can also be utilized for other purposes, such as reasoning, image recognition, and language understanding. For example, Generative Adversarial Networks (GANs) [39] can be categorized under Generative AI, but they are distinct from foundation models. Therefore, although Generative AI is a broader domain than foundation models and deep learning, some applications of foundation models extend beyond the scope of Generative AI. Consequently, since the purpose of this study is not centered around content generation, we have focused on and scoped the use of foundation models specifically for robots operating in unstructured environments.

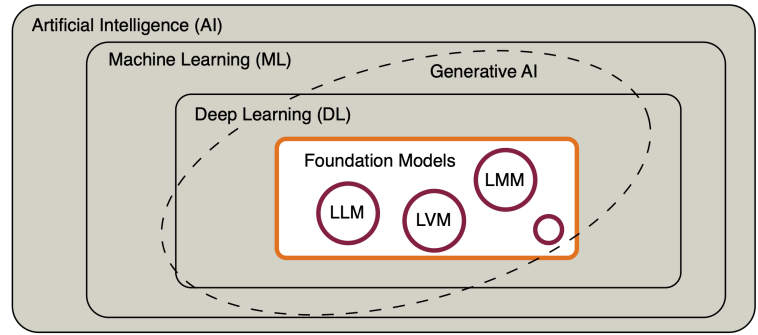

*Figure 3* The clarification of AI solution domains

### 3.3- Transformer Architecture

Three main characteristics of foundation models, in-context learning; scaling law; and homogenization, can be attributed directly and indirectly to the Transformer architecture of foundation models. The impact of the transformer architecture on creating foundation models cannot be overstated. Introduced in a study by Vaswani et al. in 2017 [40], transformers revolutionized the processing of sequential data through self-attention mechanisms, which assess the relevance of different parts of the input data, thereby facilitating in-context learning. This architecture has become the backbone of many state-of-the-art foundation models, as it supports parallel processing of sequences and effectively captures long-range dependencies within the data. The scalability of Transformers is a key factor in their success, as they can be expanded to accommodate the vast amounts of data required for training foundation models, adhering to the scaling laws characteristic of these models. Moreover, the homogenization aspect is well-served by Transformers, as their encoder and decoder components enable their application to various data types, including text, images, and audio, making them versatile tools in the development of generalized AI solutions.

Furthermore, the structure of the Transformer architecture is crucial for its performance across various tasks. This structure can be categorized into three main types: encoder-only, encoder-decoder, and decoder-only architectures [41]. For instance, BERT [42] models, which are encoder-only, excel in tasks such as question answering and text classification, where comprehending the meaning of the input text is critical. This architecture paradigm allows the model to build a better understanding of tokens and the context in which they are used. The encoder-decoder architecture is utilized in tasks such as machine translation, where an encoder comprehends the source language, and a decoder generates the target language. Models like CLIP [43] employ this structure for tasks like image captioning and text summarization. Additionally, in recent years, an increasing number of models have adopted decoder-only architectures, which have demonstrated remarkable performance [44]. These architectures are trained to generate the next word (or any token) in a sequence given the preceding words. Models such as GPTs are prime examples of these architectures and have shown reasonable few- and zero-shot performance in various tasks such as text generation.

### 3.4- Modalities

Various specific applications can be built over foundation models, and they are growing fast. To fully grasp applications of foundation models for robotics in unstructured environment, we classify them based on the modalities they are built over. The Table 2 presents the list of foundation models categorized by various modalities, along with their corresponding models for each modality. The subsequent subsections investigate further details regarding these models.

*Table 2* Foundation models based on different tasks and modalities.

| Modalities | Foundation Models | Purpose | Example Models |
|---|---|---|---|
| Text | LLMs | Achieving state-of-the-art performance on language-based tasks. | GPT-2, BERT, GPT-3, GPT-4 |
| Image | LVMs | Improving image classification, segmentation, and object detection. | ViT, ViT-G, ViT-e |
| Multimodalities | Vision-Language | Handling combined inputs of images and text for tasks like image captioning and visual question answering. | GPT-4, CLIP, BLIP, FILIP |
| | Language-Audio | Encoding sound into vector space and facilitating similarity computations for audio tasks. | AudioCLIP, CLAP, Whisper |
| | Language-Vision-3D data | Incorporating spatial information and 3D data into architectures for tasks like segmentation and object detection. | 3D-LLM, OpenScene, SpatialVLM |
| | Sensor data | Utilizing various sensor data types like IMUs, heatmaps, and skeletal movements for multimodal learning and task execution. | ImageBind, Meta-Transformer, FoundationPose, Human Motion Diffusion Model, T2M-GPT, GestureDiffCLIP, Kosmos-2 |

### 3.4.1- Text Modality: Large Language Models (LLMs)

State-of-the-art LLMs, a class of foundation models, utilize billions of parameters and are trained on a massive number of tokens, reaching into the trillions. This groundbreaking approach enabled pioneering models like GPT-2 [45] and BERT [42] to achieve remarkable state-of-the-art performance on challenging benchmarks such as the Winograd Schema Challenge [46] and the General Language Understanding Evaluation [47]. Building upon the success of these models, the next generation, including GPT-3 [12], LLaMa [48], PaLM [49], and more recent arrivals such as Claude [16] and LlaMa-2 [14], followed the same formula while substantially increasing the parameter count to over 100 billion, expanding the context window size to accommodate over 1,000 tokens, and utilizing training datasets of unprecedented magnitude, often spanning tens of terabytes of text. These models are expanding so fast that we are seeing new LLMs every few days.

A prime example of this paradigm is the highly successful GPT models, which was trained on the vast Common Crawl dataset [50], a repository containing petabytes of publicly available data amassed over 12 years of extensive web crawling, encompassing raw web page data, metadata, and extracted text. One of the remarkable capabilities of LLMs, akin to other foundation models, is their ability to undergo fine-tuning, a process wherein their parameters are adjusted using domain-specific data to align their performance with specialized use cases. Notable instances of this include GPT-3, which has been fine-tuned through reinforcement learning with human feedback (RLHF) [51], enabling it to excel in various tasks tailored to specific domains.

### 3.4.2- Image Modality: Large Vision Models (LVMs)

It is predicted that LVMs would be the next revisionary AI solution after LLMs [52]. Akin to LLMs, foundation models in the vision domain can be trained on a massive corpus of images, leveraging millions or even billions of parameters, enabling them to learn intricate patterns within visual data. Transformer (ViT) [53], [54] models are among the most significant Large Vision Models (LVMs), harnessing the power of the Transformer architecture to bring the capabilities of foundation models to computer vision tasks, such as image classification, segmentation, and object detection [55]. ViTs treat an image as a sequence of image patches, referred to as tokens. During the image tokenization process, an image is divided into patches of fixed size, which are then flattened into one-dimensional vectors, known as linear embeddings. To capture the spatial relationships between image patches, positional information is added to each token [56].

Recently, ViTs have demonstrated impressive results compared to conventional Convolutional Neural Network (CNN) models across various tasks and datasets [57]. Techniques such as efficient attention mechanisms [58], compression methods [59], and enhanced training strategies [60] have been employed to improve the performance and efficiency of ViTs. Notable examples include ViT-G [61], a scaled-up ViT model with 2 billion parameters, and ViT-e [62], boasting 4 billion parameters. Furthermore, ViT-22B [63], a vision transformer model with a staggering 22 billion parameters, has been employed in PaLM-E and PaLI-X [64], contributing to robotics tasks. Additionally, SAM [65] provides zero-shot promptable image segmentation capabilities.

### 3.4.3- Multi-modality: Large Multimodal Models (LMMs)

Multimodality refers to the ability of a model to accept different "modalities" or types of inputs, such as images, text, or audio. Vision Language Models (VLMs) are a category of multimodal models that take in both images and text for various tasks, including image captioning, visual question answering, and visual entailment. One widely used VLM is Contrastive Language-Image Pre-training (CLIP) [43], trained on 400 million image-text pair datasets from the internet. CLIP offers a method to compute the similarity between textual descriptions and images. It employs internet-scale image-text pairs data to capture the semantic information between images and text. The CLIP model architecture comprises a text encoder and an image encoder (a modified version of ViT) that are jointly trained to maximize the cosine similarity of the image and text embeddings. Another notable example is BLIP [66], which focuses on multimodal learning by jointly optimizing three objectives during pre-training: Image-Text Contrastive Loss, Image-Text Matching Loss, and Language Modeling Loss. This method leverages noisy web data by bootstrapping captions, enhancing the training process. FILIP [67] concentrates on achieving finer-level alignment in multimodal learning. It incorporates a cross-modal late interaction mechanism that utilizes token-wise maximum similarity between visual and textual tokens. This mechanism guides the contrastive objective and improves the alignment between visual and textual information.

Recent advancements in multimodal foundation models, enabled them to handle a wider range of modalities and tasks. Prominent examples of such models are AudioCLIP [68], CLAP [69], and Whisper. AudioCLIP and CLAP are more focused on encoding sounds into vector space, facilitating similarity computations, while Whisper [70] seeks to specify easily confused words and perform in-context learning. Furthermore, there are foundation models that incorporate spatial information and 3D data into their architectures to enable segmentation and object detection tasks. Some of these models are 3D-LLM [71], OpenScene [72], and SpatialVLM [73]. It's worth mentioning that multimodality is not limited to image and text data; potentially, any sensor data, such as IMUs, heatmaps, object poses, and skeletal movements, can be used to create multimodal foundation models. Notable models in this regard include ImageBind [74] and Meta-Transformer [75]. These methods, similar to CLIP, and CLAP, enable similarity calculations but can simultaneously handle a significantly larger number of modalities.

Additionally, FoundationPose [76] is a unified foundation model for 6D object pose estimation and tracking, supporting both model-based and model-free setups. There are numerous techniques for dealing with gestures and skeletal movements, such as the Human Motion Diffusion Model [77], T2M-GPT [78], and GestureDiffCLIP [79], which generate human motion or gestures through spoken language. More recently, models like Kosmos-2 [80], have pushed the boundaries of multimodal learning by successfully tackling a vast array of tasks across different modalities, including vision, language, speech, and robotics, using a single unified architecture.

### 3.5- Related Studies

To date of drafting this manuscript, only four studies have reviewed the current application of these models in robotics. The first review paper by Firoozi et al. [81] surveyed the application of foundation models in robotics with an emphasis on future challenges and opportunities, including safety and risk. The second review paper by Xiao et al. [19] explored existing studies focused on robot learning using foundation models to identify potential future areas. In the third review, Hu et al. [82] examined different studies relevant to foundation models and investigated how their application could be adapted to the robotics field. The fourth study [31] emphasized the replacement of some current components in robots with foundation models.

Despite the invaluable contributions of these studies, they lack a transparent and reproducible approach in their research method and toward their findings. Their method fails in quantitively highlight the significance of each application, making it difficult to compare and analyze which applications need more attention from researchers. Their method also falls short in identifying subtle research gaps that are not apparent through narrative literature review. Consequently, there is a lack of an objective benchmark in the field to track progress and ensure that studies are advancing safely and aligning with our goals.

This study is distinguished from previous ones for the following reasons: (1) This study focuses on unstructured environments and how foundation models can enable robots to integrate into such settings; (2) To achieve study objective, this study employs a systematic, reproducible, transparent, and multi-disciplinary approach that see the problem from three different perspectives; (3) It builds an objective picture of the current state-of-the-art in employing foundation models for robotic and unstructured environment applications, which serve as a clear benchmark for future research; (4) This study synthesizes findings from the current state-of-the-art to envision future scenarios, challenges, and potential directions aimed at developing safe autonomous robotic systems for unstructured environments.

## 4. Research Method

### 4.1. Schema Overview

Our study objectives are designed in the intersection of three fields, namely foundation models, robotics, and unstructured environments that are less investigated within a single lens. As a result, this study employed a systematic multi-dimensional approach toward study objectives to not only fully grasp the current development stages in each field, but also synthesized them with theory to picture future pathways, challenges, and potential solutions. This multi-dimensional systematic approach is built over prior studies [83], [84], [85], transitioning to a multi-dimensional schema tailored for our specific literature review. This schema is designed to clarify the interrelations and dependencies between the domains under this study. Figure 4 illustrates the conceptual view of our outlined schema. The z-axis represents the Foundation Models domain, the x-axis encapsulates the Robotics domain, and the y-axis symbolizes the Unstructured Environment, illustrating how these domains interact and influence each other within the context of this study.

The schema is structured based on the intersection of planes that result from each pair of axes. The xy-plane reflects the deliberative acting theory, and desired characteristics of autonomy and generalizability as fundamental needs for successful adoption of robots in unstructured environments. The xz-plane explore the use of Foundation Models for Robotics, showcasing the current state of development, the areas of robotics most influenced by these models, research gaps, and emerging possibilities. Simultaneously, the yz-plane examines how foundation models are utilized for a variety of applications in unstructured environment. To achieve the exploration objectives for xz- and yz-planes, we conducted a systematic review approach, which is more detailed in next sections. By fusing these thematic planes into a cohesive xyz area, our schema offers a comprehensive analysis, facilitating a synthesis of the multifaceted insights we gather. This synthesis not only maps the existing knowledge and landscape but also forecasts potential pathways within the convergence of these thematic planes. The following sections describe our methodology for conducting systematic review over xy- and xz-thematic planes.

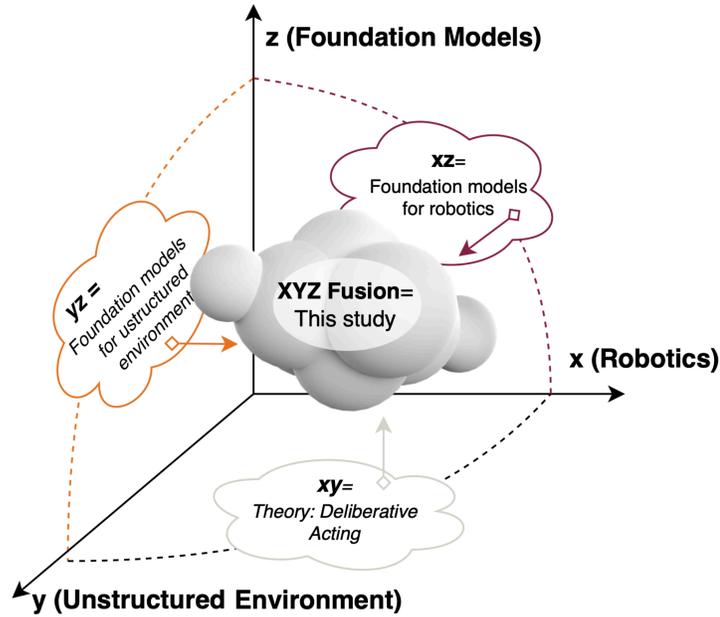

*Figure 4* Schematic view of proposed framework.

### 4.2. Database and search query

A systematic review approach has been selected for this study to explore the application of foundation models for robots in unstructured environments (represented on the xz- and yz-planes). An important factor influencing the outcome of systematic reviews is the appropriate selection of search systems, which enable an objective and repeatable review process [86]. This is especially important when the systematic review is about a newly emerged topic, such as foundation models in robotics.

Web of Science (WoS) and Scopus are two comprehensive databases serving as major tools for systematic review in the field of engineering [87], [88]. In addition to these, ArXiv is selected as the main source for preprint studies within the scope of our study because: (1) it serves as one of the main sources of studies related to foundation models from 2018 up to now [89]; (2) it helps us to cover emerging ideas that are not yet published in journals due to the long process of publishing [90]. Although Google Scholar is one of the other common search databases, studies recognize it as an inappropriate database for systematic review [89]. Through this study, we combined the search results from WoS, Scopus, and ArXiv as this approach is tested in prior systematic review efforts, leading to more comprehensive output [91], [92]. Query-based search is one of the most fundamental methods for identifying relevant studies in a field of research [93]. To maximize the potential of identifying relevant studies within our scope, we constructed two search queries (see Figure 5): one representing studies on the application of foundation models in robotics, and the other exploring the use of foundation models in unstructured environments.

Figure 5 depicts the process of building search queries as the first stage in identifying relevant studies. This process is structured around three word-family blocks containing keywords relevant to our targeted studies. Within these blocks, keywords are connected with "OR" operators to maximize the likelihood of retrieving relevant studies. Among these blocks, the word-family block for foundation models is shared with two other blocks: the word-family block for unstructured environments (left block) and the word-family block for robotics (right block). Integrating our shared block (middle block) with either side block (right or left) using the "AND" operator results in building a search query for that specific thematic domain. Linking the middle block with the left block generates a search query suitable for exploring the application of foundation models in unstructured environments (yz-plane). Conversely, connecting the middle block with the right block formulates a search query that will enable us to identify studies related to the application of foundation models in robotics (xz-plane).

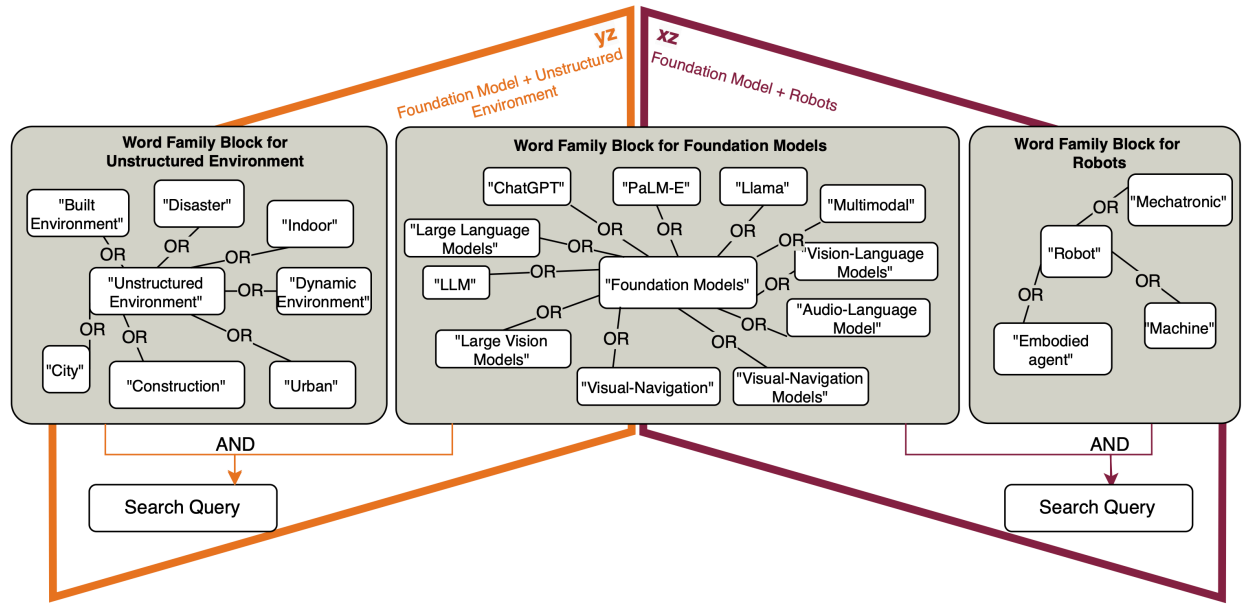

*Figure 5* The process of creating search queries for YZ-thematic plane (left); and XZ-thematic plane (right).

### 4.3. Study identification

This stage involves using the outcomes of the search query from the previous section to identify studies relevant to the scope of this study. Figure 6 illustrates the process of identifying relevant studies for foundation models in robotics (xz-thematic plane) and foundation models in unstructured environment (yz-thematic plane). Although the general process is the same for both thematic planes, different search queries have resulted in varying numbers of studies at each step. It should be noted that the number of studies at each step is dependent on the date of drafting this manuscript. Initially, the search query was applied to identified databases, followed by the exclusion of duplicate studies. Subsequently, several refinements were made to the search outputs to align them more closely with the study's scope. English was selected as the sole acceptable language for this review. Noting that the first versions of foundation models emerged in 2018, we restricted the identified studies to the time frame of 2018 to 2024. Additionally, it is widely acknowledged that many significant studies in the field of computer science are published as conference proceedings [94]. Therefore, we limited the document types to journal articles and conference proceedings to ensure a comprehensive and generalizable output. This refinement process enabled us to exclude unrelated studies.

In the next step, we established a set of exclusion criteria to ensure that the identified studies are relevant to our scope. Many of the identified studies were more relevant to the applications of education and medical science, which cannot be classified as unstructured environments and must be excluded. In the realm of foundation models in unstructured environments (yz-plane), our search query detected numerous studies containing both the terms "foundation" and "construction." However, in these cases, "foundation" was used in its civil engineering context as a type of structure rather than referring to an AI model. Consequently, these studies were excluded from our results. Moreover, since our scope is confined to unstructured environments, we also excluded papers from other unstructured environments such as agriculture and underwater studies. Furthermore, in the realm of foundation models for robotics (xz-plane), we excluded papers that did not link their findings to robotics and were more focused on exploring the capabilities of foundation models, such as ontology learning and reasoning skills in LLMs. These criteria underwent two rounds of screening, initially focusing on study titles and abstracts. In many instances, the eligibility of studies could not be determined through title and abstract checks alone, necessitating a second round of screening involving the entire text to refine our outputs. Eventually, 71 and 76 (total 147) studies remained within the scope of the study, serving as the foundation for our further analyses.

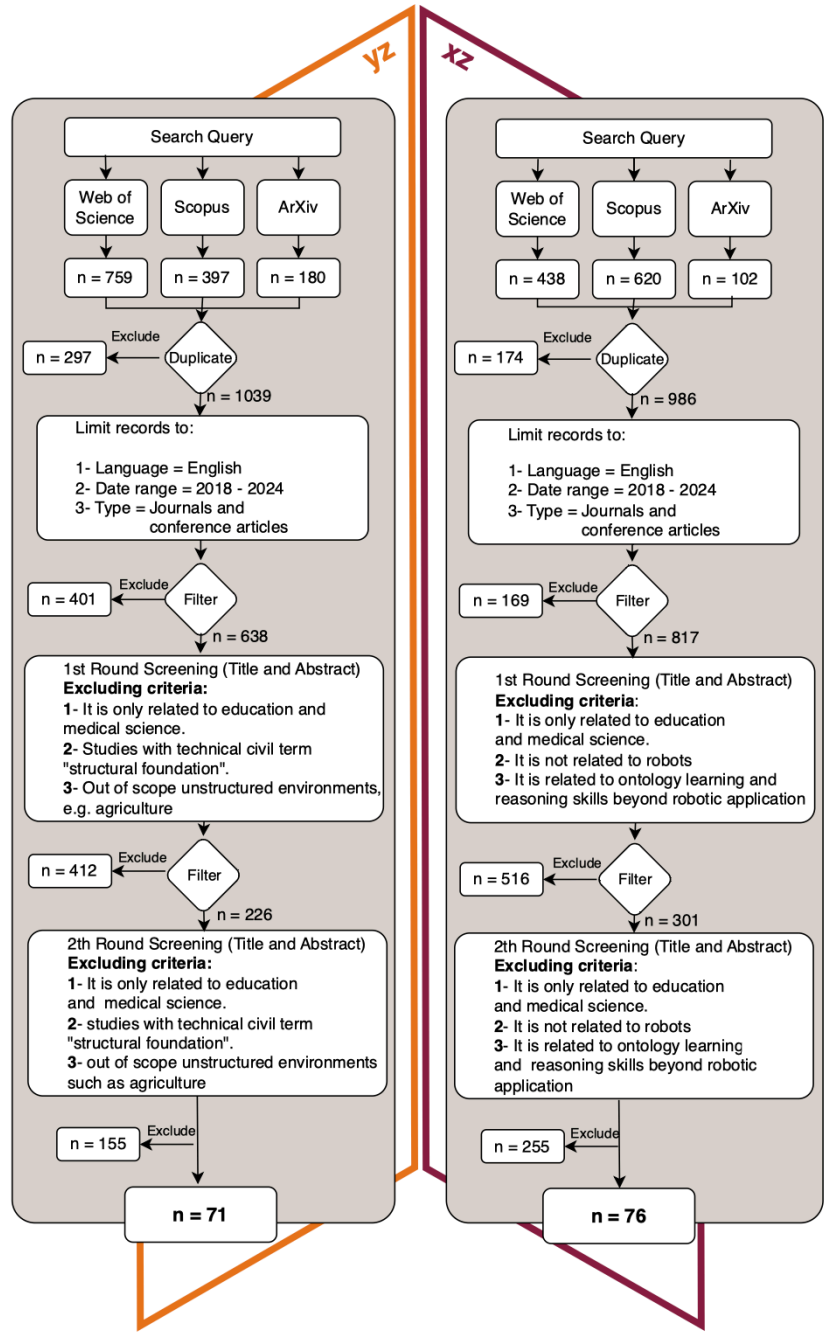

*Figure 6* The process of identifying relevant studies for xz- and yz-thematic planes.

## 5. Findings

This section aims to provide an objective picture of the current state-of-the-art in the use of foundation models for robotic and unstructured environment tasks. To achieve study objectives, all identified studies were subjected to a comprehensive whole-text content analysis. We extracted a set of 20 features to have a detailed and complete overview of the recent trends. These features can be categorized to two 10 feature groups: (1) general features:

authors, title, published year, source title, DOI, link to paper, author affiliations, abstract, author and index keywords; (2) specific features: applications, foundation model use, applied tasks, domain, study objective, robot morphology, evaluation method, modalities, transformer architecture, and open-source status. We present a summary and synthesis of the key features in this paper, with a full description and details of all features available in the project repository under an open-source license[1].

### 5.1. XZ Thematic Plane: Application of Foundation Models in Robots

### 5.1.1- Foundation model usage trends

This section investigates the frequency of utilizing foundation models for robotic applications. As seen in Figure 7, the integration of foundation models into robotics is dominated by GPT-Based models, which account for over 44% of foundation model usage. GPT-3.5 or more broadly ChatGPT, is the most frequently used LLM, highlighting its applicability and ease of use. Although GPT-4 is located in third usage place, it should be noted that the usage of GPT-4 is rapidly growing. The considerable usage between GPT-3.5 and GPT-4 can be attributed to the fact that GPT-4 is newer version and much more expensive to incorporate in robotic settings.

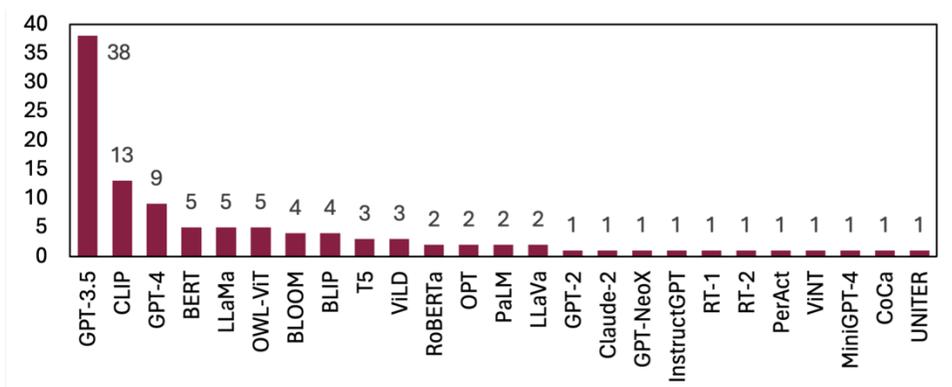

*Figure 7* Frequency of using different foundation models in robotic field.

Another interesting finding is that CLIP model is the most frequent used models among VLMs and second place among all foundation model usages in robotic applications. CLIP is primarily utilized for bridging similarities between text, as the first source of receiving language instructions, and images, as the primary means of understanding environments. It has been used in various studies to perform Vision-Language Navigation (VLN) related tasks [95], [96], as well as other manipulation tasks [97], [98], [99], and high-level recognition tasks, such as reasoning [100]. As seen in Figure 7, the frequency of remaining models is five or fewer, indicating that exploring foundation model application for robots in its early stages and more diverse models should be compared by researchers to clarify different capabilities. Considering that recent studies that highlight different biases in GPT-3.5 models [101], [102], [103], further research is needed to investigate the biases of using single models or a combination of different agents in more sensitive tasks, such as human-robot interaction or decision-making.

---
[1] Will be publicly available in the final version.

**5.1.2- Robot Morphology**

Within the identified studies of this thematic plane, 36 studies validated their findings through implementation in real robots. Single Arms are the most frequently tested robot morphology to evaluate the capabilities of foundation models, as most studies aim to validate a limited number of tasks and capabilities. The simpler morphology of a Single Arm is well-suited for this purpose. Conversely, more advanced studies using mobile manipulators with end-to-end execution are employed to cover cognitive and acting tasks without delving into more complex morphologies. All of this confirms our finding in the previous section, indicating that research in this domain is in its early stages. As a result, further studies are needed to investigate the use of foundation models in more complex tasks that require specialized morphologies such as quadrupeds, humanoids, hexapods, and soft robots.

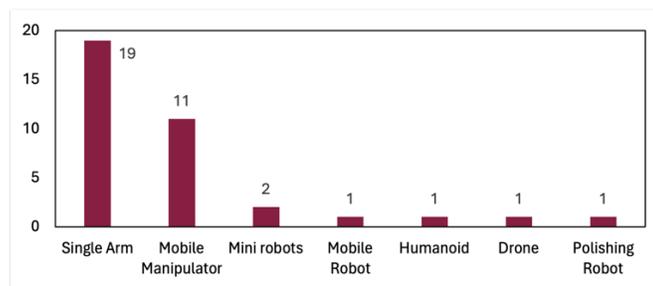

*Figure 8* Frequency of robot morphologies in foundation models

**5.1.3- Current State-of-the-art: Application of Foundation Models in Robotic Tasks**

This section aims to map the current state of the use of different foundation models in robotic tasks. To achieve this, the identified studies were categorized based on the foundation models used and the specific tasks to which these models were applied. Figure 9 illustrates the flow of applying different foundation models for robotic tasks. This figure is organized across four analytical layers: foundation models, their categories, and categories of robotic tasks, and the specific robotic tasks. Notably, almost half of the integrations of foundation models with robotics are implemented using GPT-family models.

**Foundation model categories**: Within the foundation model categories, LLMs contributed to 69% of foundation models utilized for robotic applications, indicating that most studies are exploring the capabilities of this category for addressing classic challenges in robotic domains. For example, some studies utilize the capabilities of these foundation models in understanding language and coding to generate robotics execution codes in industrial settings [104], [105]. VLMs also contributed another 20% of foundation model applications in robotic tasks, helping to bridge language instructions with vision perception in various studies [106]. However, less attention has been given to the application of LVMs in the robotic domain, where further studies are needed to analyze the applicability of these models, especially in tasks compared to traditional computer vision techniques. Recently, a few studies have gone beyond text or image-based foundation models by creating robot transformers [107], [108], yet more studies, such as MiniGPT-3D [109], are felt necessary to build 3D foundation models as they can contribute more significantly to robot-specific tasks that require direct interaction with the 3D world.

**Planning and perception tasks**: When it comes to robotic tasks, foundation models are primarily (72%) utilized for cognitive tasks rather than acting tasks (28%). One reason for this is that the current state of foundation models is better suited to cognitive tasks rather than acting, which requires extensive direct interaction with the 3D world. Furthermore, traditional robotic field have more open issues and challenges in the cognitive tasks' domain than in acting, which are historically more investigated. Within the cognitive domain, perception and planning are most common goal of using foundation models in many identified studies. For example, studies utilized capabilities of ChatGPT in understanding text to change the traditional method of robot planning, by generating behavior-tree [110] or considering the current state of robots in plan generation [111]. Furthermore, some studies focused on providing robots with better perception by utilizing foundation models in complex robotic tasks, such as scene anomaly detection [112].

**Human-Robot Interaction (HRI)** is another important use case of foundation models in the robotics field. The use of foundation models in HRI can be categorized into three main streams. First, some studies utilize LLMs to improve HRI through better extraction of machine-understandable information from human instructions [113], [114]. Another group of studies uses foundation models to understand public perceptions toward robots [115], [116], [117]. The last group applies the capabilities of LLMs to generate human-like text to respond to humans and improve trust between humans and robots [118], [119], [120].

**Reasoning and decision-making tasks**: In terms of reasoning, one mainstream application is the use of foundation models for providing commonsense knowledge to robots [121], [122]. Commonsense reasoning is a hard task for machines, but it is crucial in many tasks. For example, one study [123] utilized LLM commonsense reasoning capabilities in the field of question answering (QA), which is one of the most important tasks of NLP. Another study [124] found that LLMs are not sufficient on their own to provide commonsense reasoning but they are effective in synergy with formal knowledge representations. On the other hand, few studies investigate the decision-making abilities of LLMs in connection with different robotic tasks, such as planning [125] and manipulation [126].

**Control, manipulation, and navigation**: Beyond cognitive tasks, the capabilities of foundation models in acting tasks are less explored. For example, some studies use language understanding of LLMs as a translation module between human and robot for controlling the simple motion of robots [127], [128]. Some other studies are providing innovative frameworks for improving spatial reasoning required for robotic manipulation tasks using LLMs [129], [130]. Navigation is another challenging task that recent foundation models are used to allow researchers to have semantic reasoning and go beyond conventional map-based systems [131]. Despite these examples, acting tasks are usually come with other cognitive tasks such as planning, and perception. As a result, a network of connection between these tasks helps us to achieve better interpretation of foundation model capabilities.

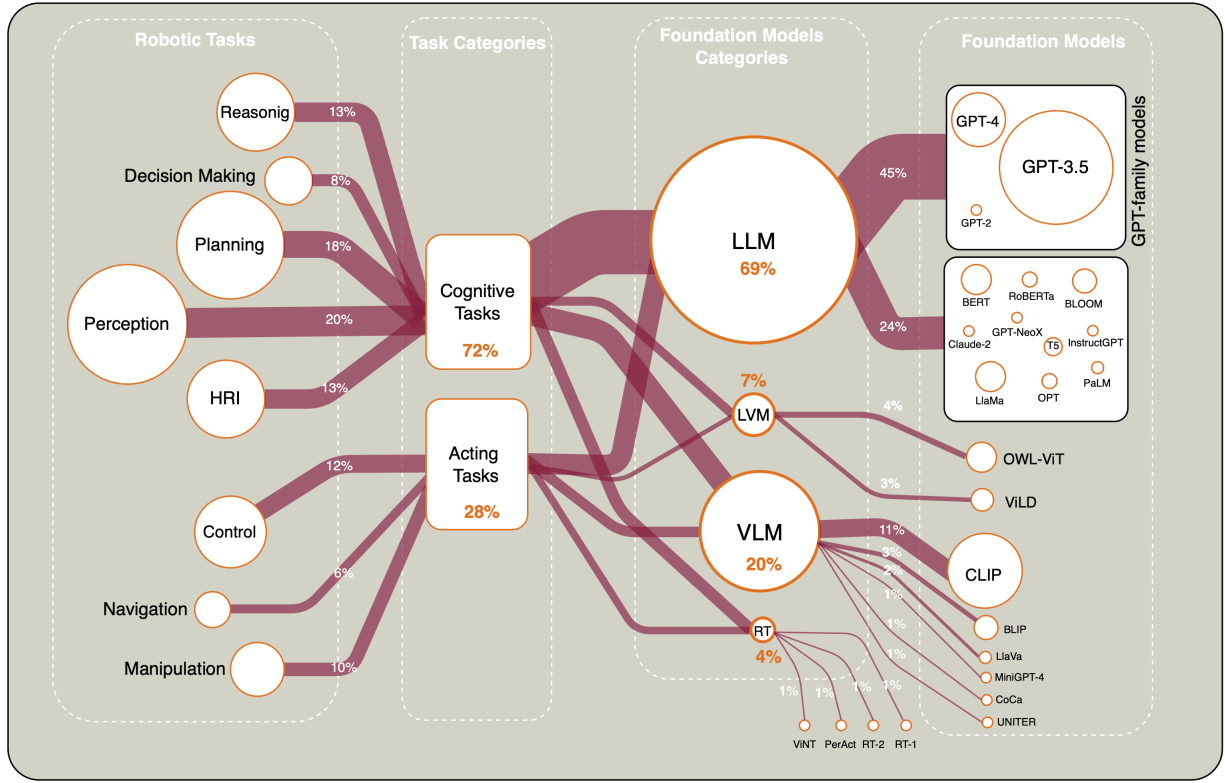

*Figure 9* Flow diagram of foundation model applications in robotic tasks

### 5.1.4- Robotic Task Integration

Robotic cognitive and acting tasks are utilized in studies in an interconnected manner to automate specific tasks. Accordingly, a network diagram can help to better demonestrate how identified studies use foundation models in different tasks. Figure 10 illustrates the co-occurrence network of robotic tasks, where cognitive and acting tasks are represented as nodes. The edges between nodes represent the co-occurrence of two tasks within a single study. The size of each node is proportionate to the number of its connections, indicating that larger nodes are more frequently utilized in conjunction with other tasks in studies. The thickness of the edges indicates the frequency of concurrent task usage in the studies.

As illustrated in Figure 10, the most significant connection is the use of foundation models for HRI and Perception. This finding, coupled with the dominance of LLMs in foundation models, indicates that most studies leverage the text analytical capabilities of foundation models to extract both defined and undefined information for planning and controlling robot actions. This suggests a need for more studies that go a step further to explore how commonsense reasoning can aid robot decision-making in situations where information is scarce, and how robots perform tasks under such constraints. For example, can robots equipped with foundation models have a commonsense understanding of construction sites and be able to determine which part of a project needs more attention? Can they also suggest solutions for engineers in the event of material shortages or supply chain delays?

The navigation node is smaller than other nodes, indicating that navigation tasks less frequently co-occur with other robotic tasks. This suggests that the majority of the field is interested in validating the capabilities of foundation models in cognitive tasks, and some acting tasks such as control and manipulation, rather than incorporating the complexity of moving in a 3D environment into their studies. This aligns with the prevalent use of a single robot arm as the primary platform for foundation models. Another interesting finding is that all edges leading to the decision-making node are thin, which indicates that this task is also overlooked in many studies. Therefore, more research is needed in this domain to fully explore the real capabilities of foundation models. For example, robots could be tested on construction sites to assess their capabilities in navigating dynamic and ever-changing environments. Additionally, it would be valuable to explore how foundation model-equipped robots decide between several cluttered paths, some of which may have a high potential for hazards and accidents.

Despite the small size of the reasoning task node, there is a considerable connection between this node and the perception node. This represents a major category within this field, which involves utilizing reasoning capabilities to perceive situations where only a small amount of information is available, such as unseen scenes and undefined events [123], [124], [132], [133], [134]. This alignment is crucial for robots operating in unstructured environments. However, fewer studies have applied the reasoning capabilities of foundation models to acting tasks, such as navigation, control, and manipulation. This observation confirms our other findings and highlights research gaps that need addressing.

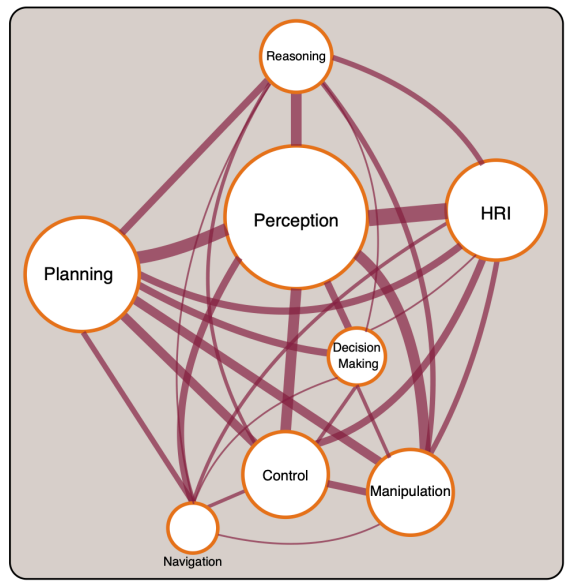

*Figure 10* Co-occurrence network of robotic tasks using foundation models.

## 5.2. YZ Thematic Plane: Application of Foundation Models in Unstructured Environments

### 5.2.1- Foundation model usage trends

This section investigates the frequency of utilizing foundation models for applications in unstructured environments. As depicted in Figure 11, similar to the observations made in Section 5.1.1, GPT-based models are the most utilized in unstructured environments. However, in these environments, GPT-4 shares equal popularity with the GPT-3.5 model, indicating that studies in this domain are more recent compared to those in the robotic domain. Furthermore, the diversity of foundation models in this domain is significantly less than in the robotic domain, suggesting that this field is still in its earliest stages of development, even more so than the robotic field. This could imply that fewer researchers are familiar with the capabilities of different models compared to those in the robotic field. Consequently, the authors recognize utilizing new foundation models or at least a diverse set of different foundation models as a way to move away from dependency on a single model and explore new features as a future direction research track. Being limited to a few models hampers the identification of new capabilities and findings in unstructured environments.

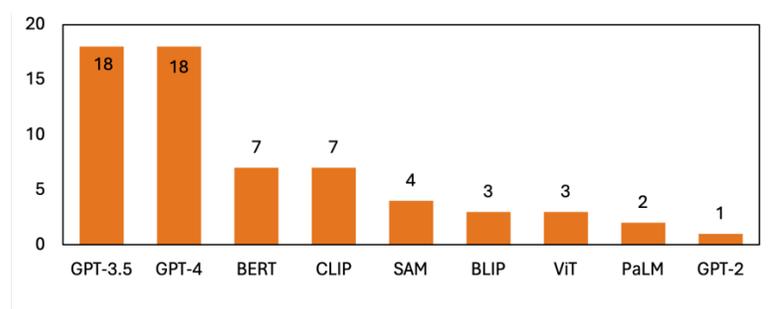

*Figure 11* Frequency of using different foundation models for unstructured environments.

## 5.2.2- Current State-of-the-art: Application of Foundation Models in Unstructured Environments

This section aims to capture the current state of the art regarding the applications of foundation models in unstructured environments through a comprehensive analysis of 71 identified studies. To facilitate this analysis, the studies were categorized into four groups: (1) the type of foundation model; (2) the category of the foundation model; (3) the specific unstructured environment; and (4) the specific applications for which foundation models are utilized in these environments. Figure 12 illustrates the utilization of foundation models across five domains of unstructured environments. The first layer from the right, representing the foundation model shows that GPT-based foundation models are the most popular and have the most diverse applications in unstructured environments which aligns with findings in the robotic field and the xz-thematic plane. This finding underscores the versatility and ease of use of OpenAI's products, highlighting their broad applicability in various settings. However, a recent study suggests that LLMs can exhibit distinct 'personalities' [135], and relying predominantly on a single model type might limit our ability to explore diverse responses, thereby constraining our ability to test the generalizability of these models.

**Foundation model categories:** Whitin the foundation model categories, LLMs, again, are dominated the field by contributing to 73% of applications in unstructured environments. For example, generative capabilities of GPT-3.5 models in creating human-like text is utilized in a study by Hussain et al. [136] to improve the human interaction through Virtual Reality (VR).

VLMs, similar to findings in xz-thematic plane, are utilized for 16% of applications in unstructured environments, with the major role of CLIP model, which for example are utilized in one study to detect changes between two remote sensing images [137]. One interesting finding is that despite the fact that application of foundation models in unstructured environment is an earliest move compared to robotic field, more researchers were interested in exploring the application of LVMs, which contributed to 11% of studies. This difference is mostly appeared in construction relevant application where ViT and SAM models are utilized to perform computer vision tasks, such as labeling indoor environments [138], and segmentation of dim and cluttered underground environments [139].

**Construction sites**: When it comes to applications in unstructured environments, the majority are related to the fields of Architecture, Engineering, and Construction (AEC), with over 37% of identified studies exploring foundation models in various construction tasks. Within this domain, construction safety is a key focus for researchers, where the capabilities of foundation models have led to improved applications such as safety training [136], hazard recognition [140], [141], question answering on incident reports [142], and worker activity recognition [143]. Moreover, foundation models have been utilized to automate project management tasks in construction, including automated reporting [144], and automated contract review [145] both using GPT-4. The application of foundation models in construction sites extends beyond these studies to include task-specific domains such as automated scheduling and planning [146], [147], integration with Building Information Models (BIMs) [148], [149], [150], and robotic assembly [151], [152].

**Disaster zones**: The significant application of foundation models in disaster zones warrants further discussion. Specific areas such as natural hazard detection and disaster reporting are more extensively explored for applying foundation model capabilities than other applications. Within hazard detection, studies utilize the visual capabilities of models like GPT-4 and Segment Anything (SAM) [153], to develop applications such as natural hazard segmentation [154], [155], [156], flood [157] and overall damage detection [158], or assessing risk in the pre-disaster stage [159]. Additionally, the linguistic capabilities of foundation models are employed in some studies to extract valuable disaster information from social media [160], [161], [162], and to automate the process of reporting in the post-disaster stage [163], [164].

**Scene and geospatial understandings**: Another significant application of foundation models is in scene and geospatial understandings, which are relevant in almost all unstructured environments. For instance, one study utilizes SAM features in segmentation to enhance scene understanding in underground sites, such as tunnels, applicable to construction and mining settings [165]. The same feature of scene understanding with minimal data is used to develop applications that caption conditions with dynamic weather changes, applicable to all outdoor unstructured environments [166]. The capability of being pre-trained on general data and possessing commonsense understanding of environments is utilized to recognize unseen objects that models were not previously trained on [167]. In terms of geospatial understanding, foundation models are employed in applications such as map assistance [168], understanding Geographic Information System (GIS) data [169], and interacting with ArcGIS tool [170], which are relevant to all outdoor unstructured environments.

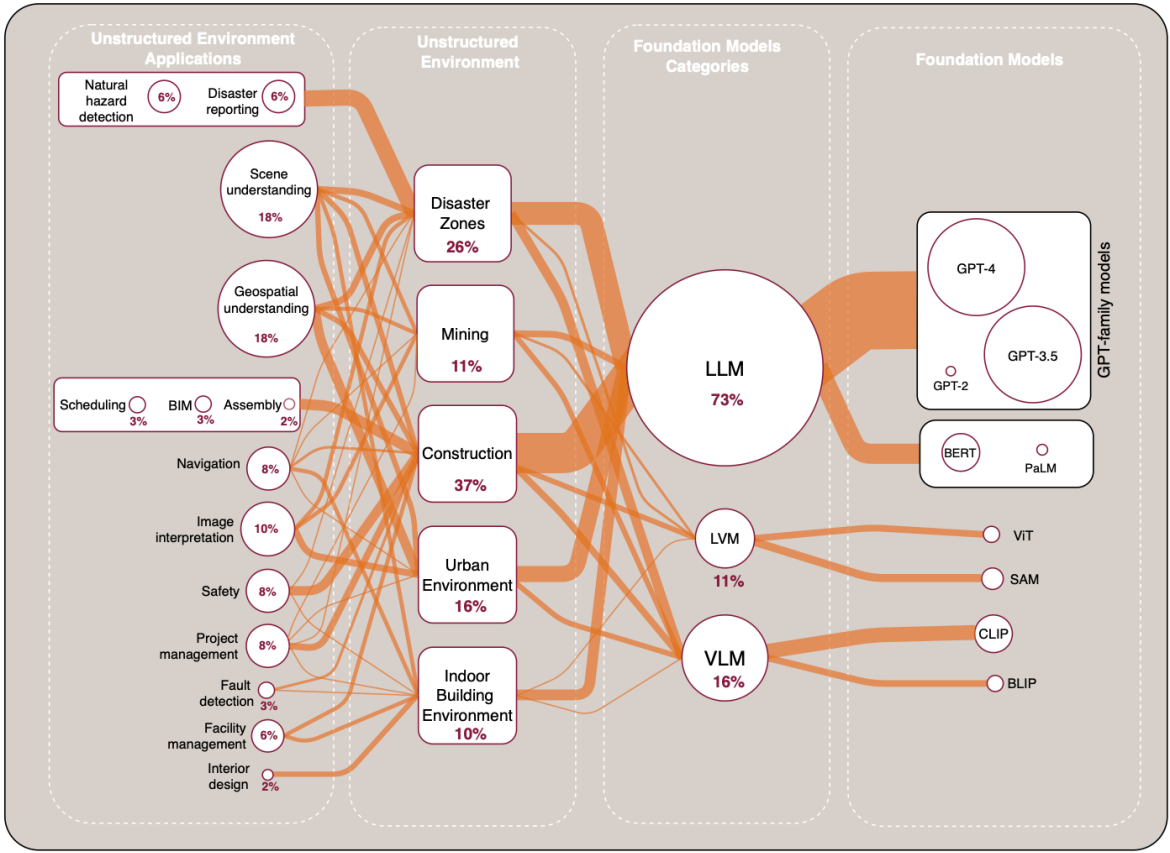

*Figure 12* Flow diagram of foundation model applications in unstructured environments

Despite the various applications mentioned in the previous paragraphs, none of the studies validated their cases by implementing robots in unstructured environments. This indicates a significant research gap in studying and testing real-world robots for tasks in unstructured settings. It also suggests that while current state-of-the-art robots are extensively tested, further steps are necessary to effectively incorporate them into unstructured environments. This gap highlights the need for more practical testing to bridge the divide between theoretical applications and real-world usability in challenging conditions.

## 6. Synthesis and Discussion

This section aims to synthesize the findings from previous sections to prospect challenges and future direction of integrating foundation models with robotics in unstructured environments. To this end, we first merged the findings in xz- and yz-thematic planes to investigate potential capabilities of robots equipped with foundation models in unstructured environments. Then, we infused findings with the deliberative acting theory.

### 6.1- Intersection Thematic Planes (xz and yz planes): Foundation models in robotics and unstructured environments

This section explores the symbiotic relationship between enhancing unstructured tasks through foundation models and their translation into robotic applications. Specifically, we aim to understand the future of robotics in unstructured environments when current identified applications are applied to them. To this end, we crossed the xz- and yz-thematic planes in Figure 13, which illustrates how tasks enabled by foundation models in unstructured environments can improve robotics in these settings. This perspective reveals the potential capabilities of robots when integrated with foundation models in these environments.

As depicted, tasks from unstructured environments are mapped onto corresponding robotic tasks, with robotic perception being notably more interconnected with other tasks. This connectivity is pivotal for future robotic applications in unstructured environments. A larger perception unit suggests that future robots could achieve enhanced cognition in the complex settings of unstructured environments utilizing foundation models. Additionally, as robots gain better perception and cognitive capabilities through foundation models, their autonomy in decision-making and reasoning could also increase. This could lead to more sophisticated decision-making processes where robots can assess risks, prioritize tasks, and make strategic decisions independently, especially in dynamic and unpredictable environments.

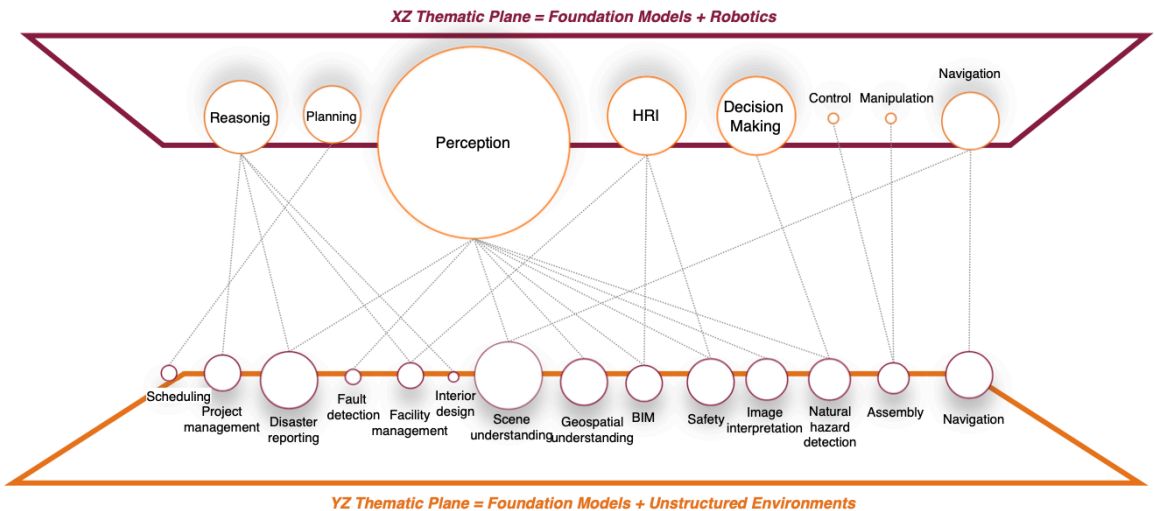

*Figure 13* Intersection of XZ- and YZ-thematic planes

For example, robots utilizing foundation models could detect natural hazards in post-disaster scenarios, analyze social media to extract more information, and then generate and reflect a more accurate disaster report, suggesting possible next actions that could save many lives. Similarly, robots could use foundation models for a deeper understanding of construction sites through schedules and Building Information Models (BIMs) to accurately assemble units while considering safety regulations, and then update project progress. The figure also indicates the potential for cross-domain knowledge transfer where foundation models trained for specific applications in one unstructured environment can be adapted to another. Studies could leverage models initially trained for purposes such as planning and reasoning and repurpose them for scheduling, spatial reasoning, hazard detection, or interaction with unstructured physical environments.

The exposure of these tasks to robotic applications suggests that foundation models are being fine-tuned for these specific scenarios, facilitating the introduction of robots into these fields in unstructured environments. However, the smaller size of control and manipulation tasks indicates that fewer studies have tested and validated the capabilities of foundation models with embodied agents in unstructured settings, confirming our discussed need for more research into the interaction of embodied foundation models in three-dimensional unstructured environments. Consequently, incorporation of robotics into unstructured environment empirically is in dire need of training and fine-tuning foundation models on acting tasks, including control, manipulation, and navigation.

## 6.2- Fusion of XYZ planes

This section integrates findings of xz- and yz-thematic planes with deliberative acting theory to (a) provide a benchmark regarding the Level of Autonomy (LoA) of future robots; (b) locate current state-of-the-art robots in LoA; (c) describe future potential pathways and scenarios and open questions.

### 6.2.1- Benchmark: Level of Automation (LoA)

It is essential to establish a benchmark to track the progress of robots equipped with foundation models. Various systems exist for classifying robots' Level of Autonomy (LoA), with the most recognized being the scale ranging from 0 (no automation) to 5 (fully autonomous), originally developed for vehicles by the Society of Automotive Engineers (SAE) [171]. Inspired by this SAE standard, Table 3 presents a newly adapted LoA scale tailored specifically for robotics, serving as a benchmark for this study. This scale assesses the independence of robots in both cognitive and acting tasks, as well as their generalizability across different tasks and the complexity of the environments in which they operate. In the table, the letters "H" and "R" represent Human and Robot, respectively, indicating their roles or involvement level within each autonomy stage.

*Table 3* Level of Autonomy for robots in unstructured environments.

| Level of Autonomy | Cognitive Tasks | Acting Tasks | General-izability | Description |
| --- | --- | --- | --- | --- |
| **Level 0:** No Automation (Manual Control) | H | H | Very low | The robot requires full human intervention for all tasks, performing no autonomous interaction with its environment. (e.g. Excavator) |
| **Level 1:** Basic Assistance | H | H-R | Very low | The robot performs some acting tasks under human supervision; cognitive tasks remain fully human-driven. (e.g. Vacuum robot) |
| **Level 2:** Partial Automation | H-R | H-R | Low | The robot autonomously handles specific cognitive and acting tasks in controlled environments, with human oversight required. (e.g. Autonomous vehicles) |
| **Level 3:** Conditional Automation | H-R | H-R | Medium | The robot independently manages tasks in specific environments, needing some human intervention. |
| **Level 4:** High Automation (Cognitive automation) | R | H-R | High | The robot is cognitively automated but the coordination between physical acting and cognitive abilities is not complete. |
| **Level 5:** Full Automation (Embodied AGI) | R | R | Very high | The robot is independent in spatial reasoning and intelligence, capable of performing tasks beyond human cognitive and acting abilities (Embodied AGI) |

### 6.2.2- Current state-of-the-art

Figure 15 illustrates the trajectory of robots and foundation models in unstructured environments. The vertical axis represents the degree of complexity in unstructured environment, and horizontal axis represent LoAs defined in the previous section. Previous findings enabled us to locate the current state-of-the-art robotic somewhere between Levels 2 and 4, where foundation models equipped robots to automate some cognitive and acting tasks and reach medium level of generalizability. The rational for this positioning is that most studies in xz-thematic plane were focused primarily on utilizing linguistic capabilities of foundation models for smoothly connecting and translating human instructions into different downstream robotic tasks. As a result, most studies are largely involved in different aspects of HRI and planning. This is also true for most studies in yz-plane, where LLMs are most explore foundation models for applications that benefit more from cognitive abilities of these models rather than spatial reasoning and intelligence. The current state of art is dominated by single modality models that usually accept text or image format data and falls short in processing 3D spatial data that is more common and most important data for interacting with environment. At this level, robots might be able to autonomously perform a medium range of tasks in moderately unstructured environments, such as controlled residential settings. However, it is important to note that this state is true for most studies while we have edge cases and studies that are close to cognitive automation of tasks. But keeping a conservative perspective to the trends in this area is vital as we do not want to overpromise current advances. As a result, we can position current state close to level 3 of LoA. In the next two sections the future pathways will be examined by discussing two extreme scenarios that would cover the range of possible pathways in the future of advancing autonomy of robotics in unstructured environments.

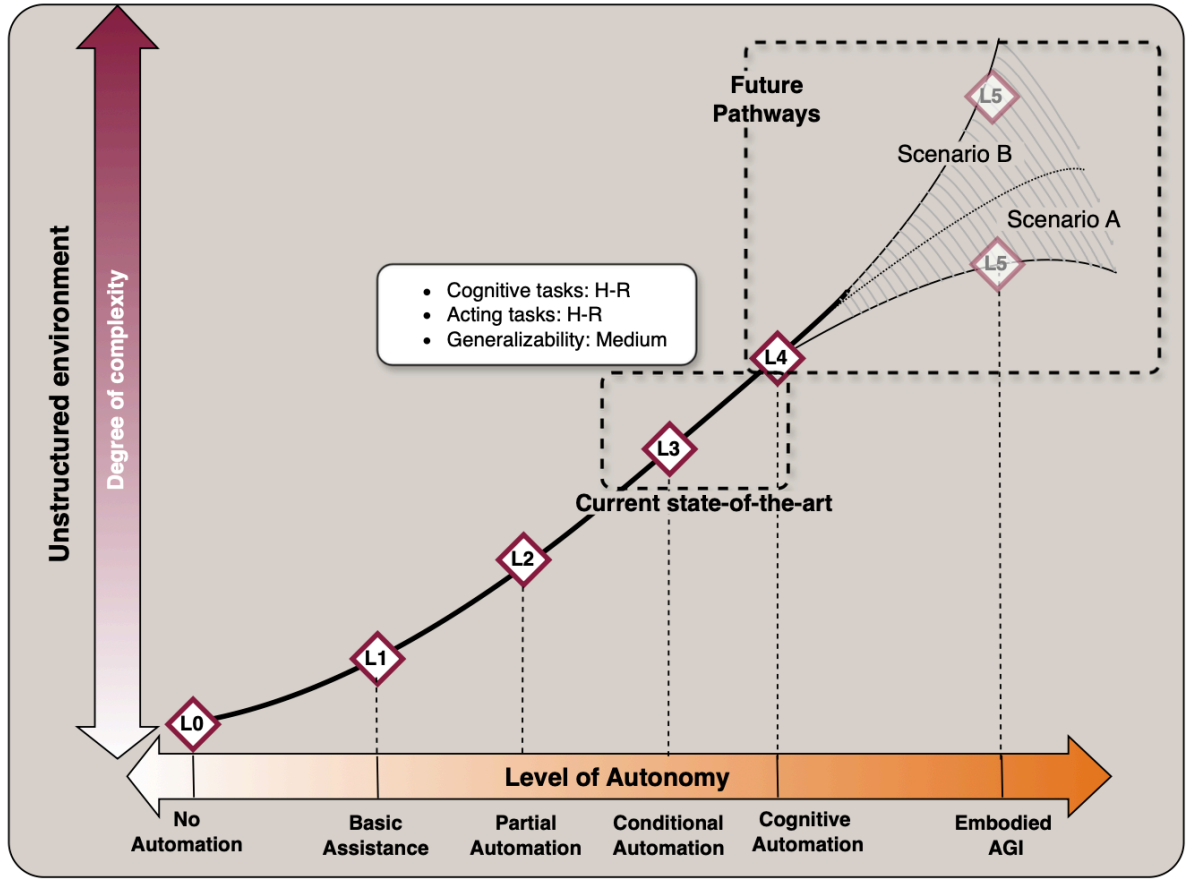

*Figure 14* Current Level of Autonomy and future pathways for robots in unstructured environments

### 6.2.3- Future pathways: Scenario A

This section aims to go beyond the current state of the art, as depicted in Figure 15, and paint the future pathways of integrating foundation models with robots for unstructured environments. Scenario A envisions a future where robots equipped with advanced foundation models will help transform naturally unstructured environments into more structured and manageable settings by: (a) reducing the degree of dynamics; (b) increasing the observability of the environment; and (c) reducing the uncertainty of unstructured environments.

In (a), robots will be equipped with advanced foundation models that can stabilize operations and maintain order. This will lead to better control over the dynamicity of the environment. In (b), foundation models will accept diverse set of data format that will enable robots to better map and analyze surroundings in real-time, providing valuable data to human or other operators. This will lead to better observability in cluttered and chaos environments. In (c), advanced robots will also help us better assess and manage risks through their analytics capabilities. This will result in a more manageable and structured environment. All of these will bring about a more manageable and structured environment, reducing the degree of complexity as a result. For instance, in a construction site, if a structural component shows signs of failure, robots can alert human workers and take preemptive actions to prevent accidents. Their ability to continuously monitor the environment and predict potential issues will significantly enhance safety and performance. Another example is when introduction of robotics and their logical work order would reduce the chaotic and cluttered manner that human crews execute tasks resulting in a more structured environment.

### 6.2.4- Future pathways: Scenario B

This section explores an alternative future pathway where the integration of foundation models with robots for unstructured environments leads to less favorable outcomes in terms of change in the degree of complexity. Scenario B envisions a future where robots, despite being equipped with advanced foundation models and other solutions, contribute to (a) increasing the degree of dynamics; (b) reducing the observability of the environment; and (c) increasing the uncertainty in unstructured environments.

In (a), robots equipped with advanced foundation models may introduce additional complexities and variability into the environment. Instead of stabilizing operations, their presence and activities might lead to more frequent and unpredictable changes. For instance, in a construction site, the movement and actions of multiple robots might interfere with each other and with human workers, leading to a more chaotic and less controlled environment. This increased dynamicity can make it difficult to maintain a stable and predictable workflow. This challenge can be identified as one of the future directions of research to enable multi-robot-human collaboration in an efficient manner.

In (b), as robots operate in unstructured environments, they may generate vast amounts of data that overwhelm human operators and existing monitoring systems. The sheer volume of information could obscure critical insights and make it harder to maintain situational awareness. For example, in a disaster response scenario, the continuous stream of data from various robots might make it challenging for responders to focus on the most important information, leading to delayed or missed critical decisions. As a result, developing AI systems that would handle such scenario can be identified as one of the future directions of research.

In (c), advanced robots might also contribute to greater uncertainty by introducing new risks and complexities that are difficult to anticipate and manage. Their interactions with the environment and with each other could result in unforeseen consequences and emergent behaviors that exacerbate the inherent unpredictability of unstructured settings. For instance, Foundation models are prone to "hallucinations," where they generate outputs that are factually incorrect or illogical. This uncertainty in the robots' decision-making processes could lead to unsafe and unpredictable actions.

### 6.2.5- Challenges and Insights: Moving Toward Scenario A

**A-Toward spatial intelligence:** One of the biggest challenges in current state of foundation model and robot integration is the lack of massive and internet size 3D data that can help robots to take another step toward spatial intelligence. This challenge is also mentioned in other studies due to the scarcity of 3D data for training foundation models [81]. Figure 14 illustrates the loop of enhancing the spatial capabilities of current robotic foundation models aimed at integration with unstructured environments. This cycle includes three main stages: implementing and interacting with virtual or real 3D unstructured environments, fine-tuning models based on synthetic data, and improving models. There are two primary approaches for integrating foundation models with 3D unstructured environments: (1) implementation in the real world; (2) simulations. Although the first option provides us with abundant valuable data and maximizes interaction with 3D worlds, there are safety challenges that hinder its complete implementation. However, any case studies in controlled and safe unstructured environments can add value and help to complete the loop.

On the other hand, simulators such as Habitat [172], VirtualHome [173], and RoboTHOR [174], have been introduced as potential solutions. However, the current state of simulators is restricted by limited situations and constraints, making them less than ideal for capturing the dynamic nature of unstructured environments. One potential solution to this challenge is utilizing LVMs, such as Sora [175], which could extend simulation capabilities and improve the execution of acting tasks in unstructured environments. Another potential solution is the use of Digital Twins, which are digital replicas of real-world environments [176]. Personalized Digital Twins of unstructured environments could allow us to implement robotic foundation models in specific settings with unique features and fine-tune robotic acting tasks based on those specific environments, such as a particular construction site or mining zone.

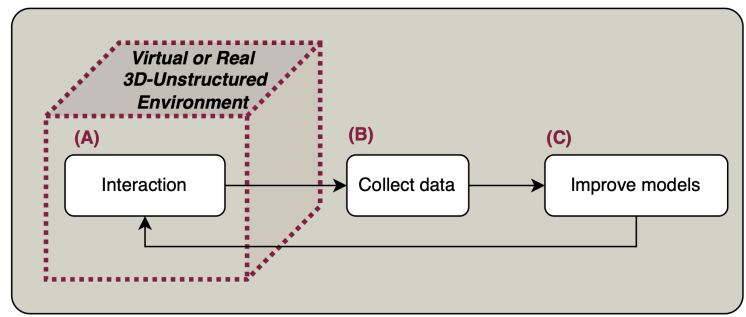

*Figure 15* Loop of preparing foundation models for unstructured environments

**B-Prevent Hallucination:** A primary issue toward Scenario A is the tendency of most foundation models to "hallucinate," meaning they often generate outputs that are factually incorrect, logically inconsistent, or physically infeasible. This problem typically arises when foundation models lack valid responses to certain situations, a scenario exacerbated by the dynamic nature of unstructured environments. This issue becomes particularly critical when robots are expected to perform a broader range of general tasks, especially those rarely encountered in their training data. Such inaccuracies are unacceptable in sensitive, human-centered unstructured environments where incorrect actions can have serious consequences. One solution to mitigate this challenge involves connecting foundation models to an updated pool of data, allowing them to adapt and respond accurately in varied situations. The Retrieval-Augmented Generation (RAG) approach in Large Language Models (LLMs) exemplifies a potential solution to this problem. Additionally, ongoing research aims to equip robots with the capability to seek assistance when uncertain about an action or cognitive task. For instance, Ren et al. [177] introduced a method called "KnowNo," which measures the cognitive uncertainty of LLMs in robotic tasks and prompts the system to request help in such situations.

**C-Improving Trust and Human Robot Collaboration**: Enhancing interaction in human-centric situations is a critical goal for autonomous robots. This necessitates a transition from the current stage, where robots are primarily command-receiver to more interactive versions, where robots generate messages and reports, to a more dynamic involvement in human-centric environments, such as urban areas. Mahadevan and et al. [178] proposed a method to generate expressive robot motions, leveraging the social context of LLMs to make robot behavior more adaptable and acceptable in social settings. However, more research is needed to develop new methods of interaction that extend beyond mere language understanding. These methods should demonstrate support and empathy, fostering improved trust between humans and robots.

Discussing trust leads naturally to the crucial role of ethics in robotics. While a comprehensive discussion of ethics goes beyond the scope of this study and requires extensive exploration of various ethical frameworks, it is essential to encourage more researchers to engage with this sensitive area. Key ethical considerations include privacy, safety, responsibility, and the moral behavior of robots, each of which warrants thorough examination. As a potential approach to addressing these issues, Zhou et al. [179] have proposed a framework that equips foundation models with the capability for moral reasoning, drawing on diverse ethical theories. Such research is an appropriate first-step as it advances the preparation of robots for deeper integration into human-centric environments, ensuring their actions are guided by sound ethical principles.

Furthermore, we need to move away from current closed-source models, such as OpenAI family models, that dominated the studies in this area. We need more open-source models that increase the diversity in experiments as well as provide better explainability and transparency in model operation mechanics, that will lead to better understanding of capabilities and shortcoming of these foundation models in future unstructured environment.

**D-Improving Latency:** Another significant barrier to deploying independent robots in highly complex and unstructured environments, such as mining and construction sites, is the fast processing time required in situations that demand quick reactions from robots. For example, a robot might need to respond immediately to a dangerous situation on a construction site. Currently, most foundation models are hosted in data centers, and robots access these models via APIs, which introduces latency and connectivity issues that can be critical in time-sensitive scenarios. Furthermore, while integrating foundation models directly onto robots through edge computing solutions, such as the Llamma model, reduces latency, it also brings high computational costs. This can limit the range and effectiveness of cognitive and acting tasks that robots can perform autonomously due to the substantial resource requirements. However, the future looks promising due to the potential for more stable internet connections through 6G and the development of more efficient and powerful foundation models, such as Mistral. These advancements could significantly mitigate current limitations, enhancing the responsiveness and functionality of robots in complex scenarios, thus fostering greater autonomy in unstructured environments.

**E- Future Structures and Improved Efficiency**: One of the significant challenges in achieving Scenario A is that current foundation models are computationally expensive, consume vast amounts of energy, and require substantial resource usage such as water in data centers. To elevate the current stage of robotics to Scenario A, robots must excel in performing a wide range of more complex and computationally intensive tasks efficiently. Achieving this requires developing more efficient models and optimizing their deployment.

To reach this stage, we need more research focused on identifying the optimal agent-structure for achieving different levels of autonomy (LoA) and generalizability in robots. One promising path toward this objective involves investigating whether it is more effective to have several specialized agents, each fine-tuned for specific cognitive or acting tasks, or to develop a single, unified model capable of performing a broad range of tasks. For instance, studies could explore the feasibility of deploying multiple smaller models (multi-agent systems), each dedicated to a particular aspect of cognitive or acting tasks. These models could then be integrated within a single robot, ensuring that the robot operates efficiently while handling diverse tasks. This approach could potentially reduce computational overhead by distributing the workload among specialized agents.

Alternatively, another approach might involve examining the structure of robot collaboration. Researchers could investigate whether it is more efficient to have a central, powerful foundation model that performs all processing tasks, with robots connected to this central model. This centralized approach could streamline processing but might introduce bottlenecks and dependencies on communication infrastructure. Conversely, a decentralized approach could be explored, where each robot operates with its own foundation model, tailored to its specific needs and capabilities. This would allow for greater autonomy and flexibility, as robots would not rely on a central processing unit. However, this approach would require ensuring that each robot's model is sufficiently robust and capable of handling complex tasks independently. Furthermore, achieving improved efficiency will also involve optimizing the hardware on which these models run. Advances in hardware design, such as energy-efficient processors and cooling technologies, can also significantly reduce the environmental impact and operational costs of running large-scale foundation models.

7. **Conclusion**

This study presents a systematic review of the application of foundation models in enhancing the autonomy of robots operating in unstructured environments. Such environments, characterized by high levels of unpredictability, dynamics, and limited observability, have historically posed significant challenges for robotic automation. The capabilities of foundation models in providing generalized solutions against unpredictable situations have been seen as a solution for addressing these challenges. To this end, this study expanded its exploration through three different lenses: the XY lens, which views the challenge of autonomous robtics from a theoretical perspective; the XZ lens, which systematically reviews 76 studies within the field of foundation models for robots; and the YZ lens, which systematically reviews 71 studies within the field of utilizing foundation models for unstructured environment tasks.

Large language models (LLMs), particularly those from the GPT-based family, are predominantly used in both robotics and tasks within unstructured environments. Cognitive activities, such as interpreting human language for robotic task planning, are extensively investigated, while tasks requiring spatial reasoning and interaction with 3D environments warrant further research attention. In unstructured environments, research has focused significantly on tasks related to the construction industry and disaster zones, recognizing the potential for enhanced decision-making and reasoning tasks in these areas.

The synthesis of findings revealed a significant gap between the current applications of foundation models in robotics and their potential use in complex unstructured environments.

While foundation models have shown promise in enhancing robotic perception and planning capabilities, their integration with physical acting tasks remains limited. This is particularly evident in the lack of studies validating the use of foundation model-equipped robots in real-world unstructured settings. The analysis also highlighted the need for more diverse foundation models, as the field is currently dominated by a few models, potentially limiting the exploration of new capabilities and features specific to unstructured environments

Findings were synthesized to locate the current state-of-the-art of foundation models in robots for unstructured environments. Based on the five Levels of Autonomy (LoA) defined, current efforts predominantly fall between partial automation (L2) and cognitive automation (L4), with most applications being limited to small-scale cases. Despite the progress so far, reaching full automation (L5) poses several challenges that could shape future scenarios, ranging from optimistic to pessimistic outcomes. Addressing these challenges is crucial for advancing toward safer, more efficient, and higher-performance robotic operations in unstructured environments. Future research should aim to navigate these barriers, paving the way for robust and fully autonomous robotic systems.

Furthermore, the study identified critical challenges that need to be addressed to advance the field. These include the development of spatial intelligence in foundation models, which is crucial for effective navigation and manipulation in 3D environments. The issue of hallucination in foundation models was recognized as a significant barrier, especially in safety-critical unstructured environments. Additionally, the need for improved human-robot collaboration and trust, reduced latency in model processing, and more efficient model structures were identified as key areas for future research. These findings underscore the importance of a multifaceted approach to developing truly autonomous robots capable of operating effectively in complex, unstructured environments.

The study also explored two potential future scenarios as our future pathway toward full automation. Scenario A pictures a positive outcome where robots equipped with advanced foundation models help transform naturally unstructured environments into more structured and manageable settings. This is achieved by reducing the degree of dynamics, increasing environmental observability, and reducing uncertainty. In contrast, Scenario B presents a less favorable outcome where the integration leads to increased complexity, reduced observability, and greater uncertainty in unstructured environments. These scenarios highlight the potential impacts of foundation model-equipped robots on unstructured environments and underscore the importance of careful development and implementation to achieve desired outcomes

Despite the contributions of this study, as discussed before, all research studies have limitations, and the present attempt is no exception to this rule. The review process only considered studies in English and used a particular set of keywords for searching. Additionally, the screening process of core studies can be considered subjective in nature, although the process was performed three separate times to minimize error. Moreover, all analyses are based on data retrieved from WoS, Scopus, and Arxiv databases. Therefore, the findings may not fully reflect the entire available efforts and studies in the field.